\documentclass[10pt,twocolumn,letterpaper]{article}

\usepackage[pagenumbers]{cvpr} % To force page numbers, e.g. for an arXiv version

\usepackage[dvipsnames]{xcolor}

\newcommand\blfootnote[1]{%
	\begingroup
	\renewcommand\thefootnote{}\footnote{#1}%
	\addtocounter{footnote}{-1}%
	\endgroup
}
\usepackage{bm}
\usepackage{makecell}
\usepackage[accsupp]{axessibility} % Improves PDF readability for those with disabilities.

\definecolor{cvprblue}{rgb}{0.21,0.49,0.74}
\usepackage[pagebackref,breaklinks,colorlinks,citecolor=cvprblue]{hyperref}
\usepackage{xcolor}

\def\verts{\mathbf{v}}
\def\hashTable{\mathbf{H}}
\def\vertFeat{\mathbf{f}}
\def\poseCode{\boldsymbol{\theta}}
\def\shapeCode{\boldsymbol{\psi}}
\def\queryPoint{\mathbf{q}}
\def\hashTableEmbedding{\mathbf{h}}
\def\cameraDir{\mathbf{d}}
\def\cameraRay{\mathbf{r}}
\def\jKNN{k}

\def\paperID{9485} % *** Enter the Paper ID here
\def\confName{CVPR}
\def\confYear{2024}

\title{Efficient 3D Implicit Head Avatar with Mesh-anchored Hash Table Blendshapes}

\author{Ziqian Bai$^{1,2*}$ \quad
Feitong Tan$^{1}$ \quad
Sean Fanello$^{1}$ \quad
Rohit Pandey$^{1}$ \\
Mingsong Dou$^{1}$ \quad
Shichen Liu$^{1}$ \quad
Ping Tan$^{3}$ \quad
Yinda Zhang$^{1}$ \\
\normalsize{$^{1}$ Google \qquad $^{2}$ Simon Fraser University  \qquad $^{3}$ The Hong Kong University of Science and Technology}
}

\begin{document}
% \maketitle
\twocolumn[{%
\renewcommand\twocolumn[1][]{#1}%
% \vspace{-4em}
\maketitle
% \vspace{-10mm}
\begin{center}
    \centering
    \includegraphics[width=1.0\linewidth, trim={0 0 0 0}, clip]{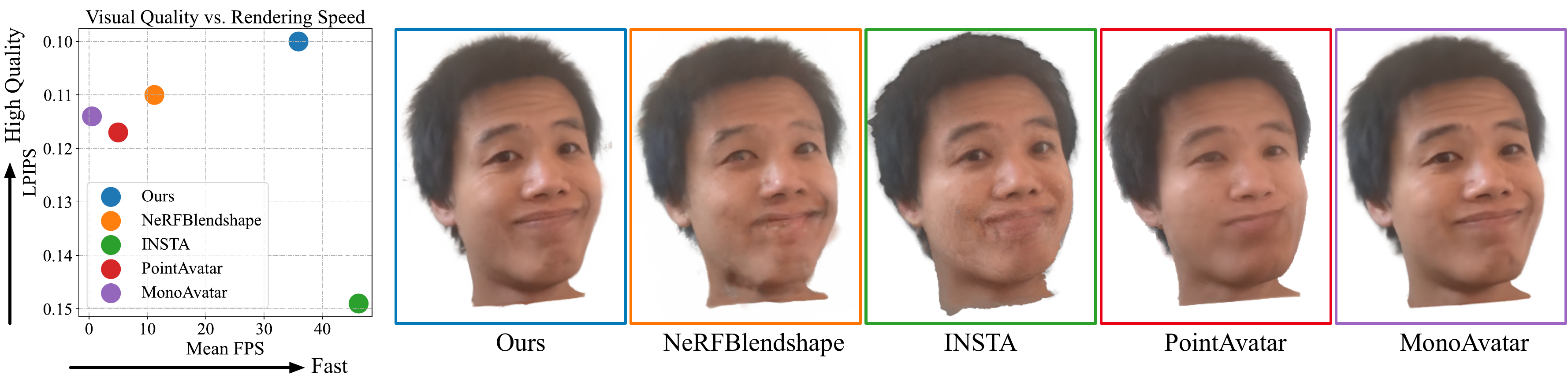}
    \vspace{-7.5mm}
    \captionof{figure}{
    Visual Quality vs. Rendering Speed comparisons between our monocular 3D head avatars and prior state-of-the-art methods including NeRFBlendshape~\cite{Gao2022nerfblendshape}, INSTA~\cite{zielonka2023instant}, PointAvatar~\cite{zheng2023pointavatar} and MonoAvatar~\cite{bai2023learning}. Our approach achieves real-time rendering (\ie $> 30$ mean FPS) with a $512 \times 512$ resolution, while produces comparable visual quality comparing to prior SOTAs~\cite{bai2023learning}, and gives significantly better results on challenging expressions than prior efficient avatars~\cite{Gao2022nerfblendshape,zielonka2023instant,zheng2023pointavatar}. Webpage \href{https://augmentedperception.github.io/monoavatar-plus/}{augmentedperception.github.io/monoavatar-plus}
    }
    \vspace{-0.075em}
    \label{fig:teaser}
\end{center}%
}]

\begin{abstract}
3D head avatars built with neural implicit volumetric representations have achieved unprecedented levels of photorealism. However, the computational cost of these methods remains a significant barrier to their widespread adoption, particularly in real-time applications such as virtual reality and teleconferencing.
While attempts have been made to develop fast neural rendering approaches for static scenes%, \eg instant NGPs~\cite{mueller2022instant}
, these methods cannot be simply employed to support realistic facial expressions, such as in the case of a dynamic facial performance.
To address these challenges, we propose a novel fast 3D neural implicit head avatar model that achieves real-time rendering while maintaining fine-grained controllability and high rendering quality.
Our key idea lies in the introduction of local hash table blendshapes, which are learned and attached to the vertices of an underlying face parametric model.
These per-vertex hash-tables are linearly merged with weights predicted via a CNN, resulting in expression dependent embeddings. Our novel representation enables efficient density and color predictions using a lightweight MLP, which is further accelerated by a hierarchical nearest neighbor search method. 
Extensive experiments show that our approach runs in real-time while achieving comparable rendering quality to state-of-the-arts and decent results on challenging expressions. 
\end{abstract}
\section{Introduction}

\blfootnote{$^{*}$Work was conducted while Ziqian Bai was an intern at Google.}

The demand of high performing photo-realistic human avatars has dramatically increased with emerging VR/AR applications, \eg VR gaming~\cite{VRChat2023,Waggoner2009My}, virtual assistant~\cite{XHoloVirtual2023}, tele-presence \cite{Orts-Escolano2016Holoportation}, and 3D videos~\cite{Guo2019The,Meka:2020}.
% \rohit{perhaps add some examples}.
How to build efficient high quality avatars from monocular RGB videos becomes a promising direction due to the convenience of monocular data acquisition.
While early works mostly adopt surface-based models for convenient controllability, recent methods (\eg MonoAvatar~\cite{bai2023learning}) leverage a sophisticated pipeline to build human avatars on neural radiance fields, which delivers vivid animations as well as significantly better rendering quality, especially over challenging parts such as hairs and glasses.
% Along this direction, the recent MonoAvatar~\cite{bai2023learning} achieves good rendering qualities and vivid expression animations by proposing a 3DMM-anchored NeRF, which conditions a neural radiance field on expression-dependent local embeddings attached to the mesh vertices of a face parametric model.
On the downside, these approaches tend to be prohibitively slow, and most of the computation is consumed by the neural radiance field inference with large Multilayer Perceptrons (MLPs).

Recently, fast approaches for neural radiance fields (\eg hash encoding in Instant NGPs~\cite{mueller2022instant}) have been proposed, which are designed mostly for static scenes or pre-recorded temporal sequences.
Despite their great success, it is not straightforward to extend these approaches for human avatars, which requires real-time rendering of dynamic facial performances when controlling the avatar.
% \rohit{check sentence}.
% Gao \etal~\cite{} proposed to learn multiple instant NGPs, one for each face blendshape. These instant NGPs are combined with linear blendshape weights to render target facial expressions.
NeRFBlendshape~\cite{Gao2022nerfblendshape} address these issues by learning multiple feature hash tables, one for each face blendshape. These hash tables are linearly combined with blending weights to render the target facial expression via hash encoding.
% serve as the instant NGPs for rendering target facial expressions.
However, the expressiveness of the avatar is compromised by the global nature of the blending shapes, which cannot accurately capture vertex-level local deformations.
On the other hand, INSTA~\cite{zielonka2023instant} proposes to build the appearance model in a canonical space using a vanilla Instant NGPs~\cite{mueller2022instant} with expression codes as the additional MLP input to capture dynamic details, which is then transformed into target expression via a face parametric model (\ie 3D morphable model (3DMM)).
% to transform all expressions into a shared 3D canonical space, then adopts the vanilla Instant NGPs~\cite{mueller2022instant} in this canonical space with expression codes as the additional MLP input to capture dynamic details. 
However, the lightweight MLP used in the vanilla Instant NGPs limits their model capacity, resulting in inferior animation quality, especially on extreme expressions.
In this work, we propose a novel 3D neural avatar system that achieves efficient inference while maintaining fine-grained controllability and high fidelity quality.
To achieve this, we introduce mesh-anchored hash table blendshapes, where we attach multiple, small hash tables to each of the 3DMM mesh vertices. These hash tables act as per-vertex ``local blendshapes'' (\ie, each ``blendshape'' is controlled by one local hash table) and influence only a local region. The mesh-anchored blendshapes are linearly merged with per-vertex weights predicted by a convolutional neural network in UV space from avatar driving signals, such as expression and head rotation. This results in expression-dependent hash table embeddings which offer several advantages over a global linear combination of blendshapes. Indeed by associating hash tables with individual vertices, we enhance the expressiveness of the model, allowing for more localized and nuanced facial expressions. This contrasts with global blendshapes, which apply uniform transformations across the entire face, limiting expressiveness. 

In more detail, our model starts from 3D query points, uses hash encoding~\cite{mueller2022instant} to gather the merged hash table embeddings from $k$-nearest-neighbor vertices around the query point, and predicts the density and color via a small MLP. The hash encoding~\cite{mueller2022instant} allows us to use a very lightweight MLP to significantly reduce computation, leading to efficient inference. Additionally, the vertex-attached hash table blendshapes represent a 3DMM-anchored neural radiance field (NeRF), which can be easily controlled by the underlying 3DMM and produce high fidelity renderings as demonstrated by MonoAvatar~\cite{bai2023learning}. To further accelerate our rendering speed, we propose a hierarchical k-nearest-neighbor search method.

Our contributions are summarized as follows.
We propose a novel approach for high quality and efficient 3D neural implicit head avatars.
At the core of our model, vertex-attached local hash table blendshapes are proposed to support efficient rendering, controllability, and capturing fine-grained rendering details in dynamic facial performances.
We also design a hierarchical querying solution to speed up the $k$-nearest-neighbor search when pulling hash table embeddings from neighbor vertices.
% and reduce overall number of query points.
Extensive experiments on multiple datasets verify that we are able to speed up avatar rendering to real-time (\ie, average $\ge 30$ FPS to render a $512 \times 512$ video) while maintaining comparable rendering quality with the state-of-the-art high quality 3D avatar ~\cite{bai2023learning} and being largely superior on challenging expressions than existing efficient 3D avatars ~\cite{Gao2022nerfblendshape,zielonka2023instant,zheng2023pointavatar}.
% comparable and even better rendering quality with the state-of-the-art high quality 3D avatar \cite{bai2023learning} and produce significantly better rendering quality than existing efficient 3D avatars \cite{Gao2022nerfblendshape,zielonka2023instant} using similar computation cost.
% . \zb{may be actually superior rendering quality... Wait to add numbers and see if change.}

\section{Related Work}

Constructing photorealistic digital humans has been a extensively researched topic. Here, we focus on discussing prior work on implicit monocular head avatars and efficient rendering. We refer readers to state-of-the-art surveys~\cite{zollhofer2018state,egger20203d,tewari2022advances} for a comprehensive literature review.

\vspace{2mm}
\noindent\textbf{High Quality Head Avatar.}
Traditionally, high-quality head avatars have been achieved under expensive equipment configurations, such as camera arrays~\cite{lombardi2021mixture,beeler2011high, Orts-Escolano2016Holoportation}, depth sensors~\cite{Chen22}, and light stages~\cite{Guo2019The,Meka2020Deep}, or require laborious manual intervention~\cite{MetaHuman2022Unreal}.
Recent research efforts have focused on constructing high quality avatars from monocular RGB videos.
One typical class of approaches~\cite{gafni2021dynamic,athar2022rignerf,zheng2022avatar,xu2023avatarmav} use implicit $3$D representations (\ie, neural radiance fields (NeRFs), implicit occupancy fields) to build the head avatar, which are parameterized by Multilayer perceptrons (MLPs). Although reasonable results are obtained, their rendering quality is still unsatisfactory especially for more challenging expressions.
More recently, Bai \etal~\cite{bai2023learning} proposed a head avatar based on $3$DMM-anchored NeRFs with expression-dependent features produced by a convolution neural network in UV space. 
Chen \etal~\cite{chen2023implicit} designed local deformation fields to capture expression-dependent deformations applied on canonical NeRFs.
Despite the impressive results, their methods didn't demonstrate real-time rendering capability due to expensive inference using large MLPs. In contrast, with our proposed mesh-anchored hash table blendshapes (\cref{sec:method}), we achieve much faster rendering speed while maintaining high fidelity results.

\begin{figure*}
    \centering
    \includegraphics[width=0.995\linewidth]{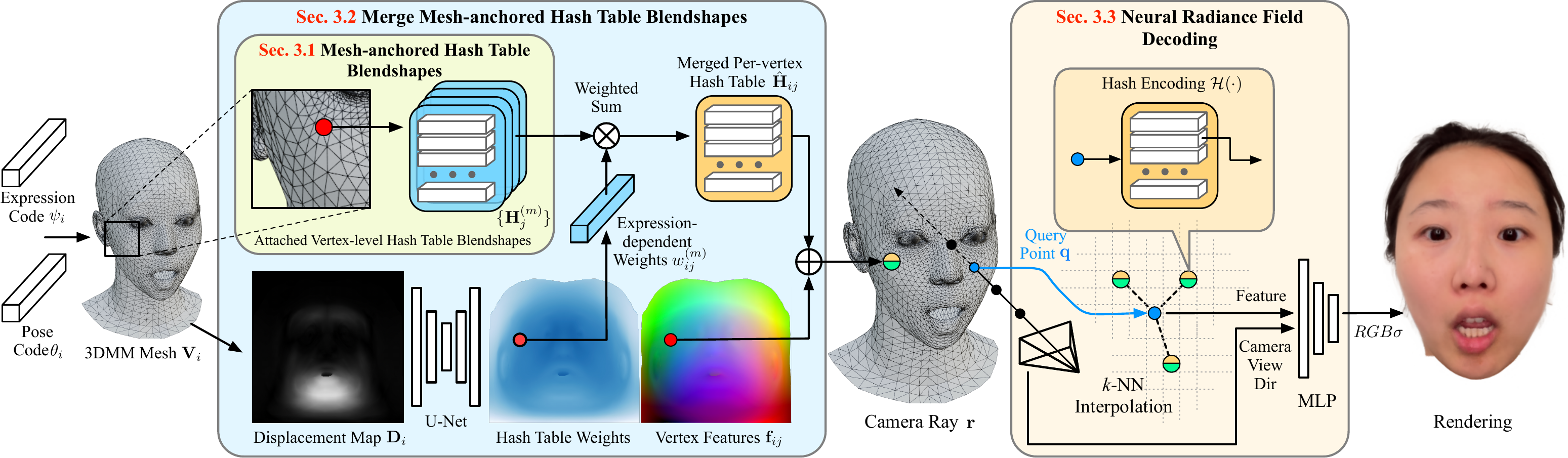}
    \vspace{-0.5em}
    \caption{Overview of our pipeline. Our core avatar representation is Mesh-anchored Hash Table Blendshapes (\cref{sec:hash_blendshape}), where multiple small hash tables are attached to each 3DMM vertex. During inference, our method starts from a displacement map encoding the facial expression, which is then fed into a U-Net to predict hash table weights and per-vertex features. The predicted weights are used to linearly combine the multiple hash tables attached on each 3DMM vertex (\cref{sec:merge_blendshape}).
    During volumetric rendering (\cref{sec:nerf_dec} and \cref{fig:nerf_dec}), for each query point, we search its $k$-nearest-neighbor vertices, then pull embeddings from the merged hash tables and concatenate with the per-vertex feature to decode local density and color via a tiny MLP with two hidden layers.
    % The core of our method is the Avatar Representation (\cref{sec:avatar_representation}. Shown as the yellow area) based on a 3DMM-anchored neural radiance field (NeRF), which are decoded from local features attached on the 3DMM vertices. Then, we use volumetric rendering to compute the output image. To predict the vertex-attached features (\cref{sec:exp_dep_feats}. Shown as the green area), we first compute the vertex displacements from the 3DMM expression and pose, then process the displacements in UV space with Convolutional Neural Networks (CNNs), and sample the obtained features back to mesh vertices. 
    % \url{https://docs.google.com/presentation/d/1BHoO9SoogzOJWRiN5VIESiq67h3yM90N_JzRq8ysVw4/edit?usp=share_link}
    }
    \vspace{-0.5em}
    \label{fig:overview}
\end{figure*}

\vspace{2mm}
\noindent\textbf{Efficient Neural Radiance Fields.}
There has been a plethora of work in recent years attempting to accelerate rendering with neural implicit representations for static objects and scenes. SNeRG~\cite{liu2020neural}, DVGO~\cite{SunSC22} and Plenoxels~\cite{yu2021plenoctrees} propose to directly optimize voxel grids of (neural or SH) features for faster performance. However, their approach still requires a large memory footprint to store per-voxel features in 3D space. KiloNeRF~\cite{reiser2021kilonerf} dramatically accelerates the original NeRF by representing the scene with thousands of tiny MLPs, however, this approach requires a complex training strategy.
% and pre-training a regular NeRF as a teacher model. 
TensoRF~\cite{chen2022tensorf} factorizes the feature grid into compact components, resulting in significantly higher memory efficiency. Concurrently, Instant NGPs~\cite{mueller2022instant} utilizes multi-resolution hashing for efficient encoding, resulting in high compactness. MobileNeRF~\cite{chen2022mobilenerf} propose to represent NeRF based on polygons, which allows leveraging a traditional polygon pipeline to enable their method to run in real-time on mobile devices.
3D Gaussian Splatting~\cite{kerbl20233d} represents a radiance field with 3D Gaussian point clouds, and leverages a point rasterization pipeline to enable fast rendering.
These methods, however, cannot be easily extended to controllable dynamic contents.
% , without trading off quality when rendering dramatic expressions and deformations. 
More recently, NeRFBlendshape~\cite{Gao2022nerfblendshape} was proposed to handle controllable expressions by learning multiple hash tables for different global blendshapes and linearly combine them with expression codes. INSTA~\cite{zielonka2023instant} transforms all expressions into a shared 3D canonical space, then adopts the vanilla Instant NGPs~\cite{mueller2022instant} conditioned on expression codes to model the head avatar. Despite the fast rendering speed achieved, their methods suffer from unsatisfactory rendering quality.
PointAvatar~\cite{zheng2023pointavatar} utilizes point clouds to represent the head avatar and uses large MLPs to predict the colors and motions of each point, leading to slow inference.
Another type of works~\cite{zhao2023havatar,xu2023latentavatar,kim2018deep,koujan2020head2head,doukas2021head2head++,wang2023styleavatar} use 2D convolution neural networks to directly synthesize images (\ie 2D neural rendering) from rasterized 3DMM meshes or low resolution feature maps generated by volumetric rendering. Despite their fast speed, the 2D CNNs may break the 3D consistency, leading to temporally unstable results especially for high frequency details. 
In contrast, our method simultaneously achieves controllability, high quality, and efficient rendering with a fully 3D representation.

% Lombardi et al.~\cite{lombardi2021mixture}
% MobileNeRF~\cite{chen2022mobilenerf}

\section{Method}
\label{sec:method}

Given a monocular RGB video, our method learns a neural radiance field (NeRF) based head avatar, which can be rendered under any specified cameras, articulated poses (\ie, neck, jaw, and eyes) and facial expressions defined by a face parametric model (\ie 3DMM). We use FLAME~\cite{FLAME:SiggraphAsia2017} as the parametric model in this work, but our method can be generalized to any other mesh-based parametric models. \cref{fig:overview} shows the overview of our method.

Our goal is to design a 3D neural implicit head avatar architecture that can simultaneously achieve high image quality, controllability, and computationally efficient rendering.
To achieve this, we propose mesh-anchored hash table blendshapes (\cref{sec:hash_blendshape}) as a novel avatar representation that can leverage both advantages from recent high-quality (\ie, 3DMM-anchored NeRF~\cite{bai2023learning}) and efficient (\ie, hash encoding~\cite{mueller2022instant}) frameworks.
More specifically, we propose to attach multiple small hash tables on each 3DMM vertex. These vertex-attached hash tables form a set of local ``blendshapes'', which will be linearly merged with predicted blending weights (\cref{sec:merge_blendshape}), and decoded into a 3DMM-anchored NeRF to support fine-grained control and high fidelity rendering. 
During NeRF decoding on a query point (\cref{sec:nerf_dec}), we pull the embeddings from the linearly merged hash tables attached on $k$-nearest-neighbor ($k$-NN) vertices from the 3DMM mesh. 
% During the NeRF decoding (\cref{sec:nerf_dec}), we first find the $k$-nearest-neighbor ($k$-NN) vertices around a query point, then leverage hash encoding~\cite{mueller2022instant} to pool the embeddings from the linearly merged hash tables attached on these $k$-NN vertices. 
Using hash encoding allows us to use a very light weight MLP (only 2 hidden layers) to predict the final densities and colors, which is the key for efficient rendering. To further accelerate our approach to real-time, we leverage the fact that close query points likely share similar $k$-NN vertices, and thus propose to group the query points into voxels and hierarchically search for $k$-NN vertices (\cref{sec:hiera_knn}). Finally, our proposed avatar representation can be trained with only monocular RGB videos without any 3D scans or multi-view data (\cref{sec:train}).

\subsection{Mesh-anchored Hash Table Blendshapes}
\label{sec:hash_blendshape}
The core of our model is an avatar representation that can represent a 3DMM-anchored neural radiance field (NeRF) while allowing us to adopt the hash encoding~\cite{mueller2022instant} technique for acceleration. 
Recent approaches manage to adopt hash encoding into head avatars with different avatar representations (\ie, global blendshapes~\cite{Gao2022nerfblendshape} and canonical NeRF~\cite{zielonka2023instant}).
In contrast, our solution is built upon the most recent 3DMM-anchored NeRF~\cite{bai2023learning}, which is superior for high quality renderings as demonstrated in our experiments (\cref{sec:compare_sota}).

We propose mesh-anchored hash table blendshapes as the new avatar representation.
Given a target expression $i$ with 3DMM pose code $\poseCode_i$ and expression code $\shapeCode_i$, we first get the deformed 3DMM mesh $\mathbf{V}_i=\mathcal{F}_\mathrm{3DMM}(\shapeCode_i, \poseCode_i)$ with $J$ vertices. 
For the $j$-th 3DMM mesh vertex $\verts_{ij}$, we attach $M$ small hash tables $\{\hashTable_{j}^{(m)}\}_M$ on it, where each hash table has multiple resolutions following instant NGPs~\cite{mueller2022instant}.
Intuitively, these hash tables form a set of vertex-level ``blendshapes'' anchored on the mesh, where each ``blendshape'' is a hash table, whose embeddings encode the information of a local radiance field around vertex $\verts_{ij}$.
% for a certain expression. 
Given a target expression to render, these hash table blendshapes are linearly summed via expression-dependent weights (\cref{sec:merge_blendshape}), such that the merged embeddings encode the fine details specific to the target expression. Simultaneously, the coarse motion of the target expression is captured by the 3DMM vertex movement, which moves the attached hash tables accordingly, and hence the corresponding local radiance field.

\subsection{Merge Mesh-anchored Blendshapes}
\label{sec:merge_blendshape}
We obtain per-vertex blending weights by running a convolution neural network (CNN) on the 3DMM deformation represented in UV atlas space.
% Similar to traditional blendshapes, we need to merge the mesh-anchored hash table blendshapes described in \cref{sec:hash_blendshape}, in order to get the final hash tables corresponding to the target expression to render.
% As mentioned in \cref{sec:hash_blendshape}, we linearly merge our hash table blendshapes with weights predicted via a convolution neural network (CNN) running in UV space from the 3DMM deformation.
Specifically, 
% given a target expression $i$ with 3DMM pose code $\poseCode_i$ and expression code $\shapeCode_i$ to render, we first get the deformed 3DMM mesh $\bm{V}_i=(\bm{\psi}_i, \bm{\theta}_i)$ and 
we calculate the vertex displacements with respect to the neutral face $\mathbf{D}_i = \mathcal{F}_\mathrm{3DMM}(\bm{\psi}_i, \bm{\theta}_i) - \mathcal{F}_\mathrm{3DMM}(\bm{0}, \bm{0})$.
% To obtain the combining weights for multiple local hash tables, t
The displacements are then warped into the UV space and fed into a U-Net to predict a weights map in $\mathbb{R}^{\mathrm{H_t} \times \mathrm{W_t} \times \mathrm{M}}$, where $\mathrm{H_t} \times \mathrm{W_t}$ is the UV resolution, and $M$ is the number of hash table blendshapes on each vertex (pre-defined as 5 in our experiments).
The weights map is then sampled back to 3DMM vertices, serving as the expression-dependent weights $\{w_{ij}^{(m)}\}_M$ to take a weighted sum of the embeddings in the hash tables on each vertex, which produces the merged hash tables
\begin{equation}
    \hat{\hashTable}_{ij} = \sum_{m=1}^M w_{ij}^{(m)} \hashTable_{j}^{(m)}.
\end{equation}
The U-Net also produces a UV feature map.
We sample a per-vertex feature $\vertFeat_{ij}$ from it, similar to MonoAvatar~\cite{bai2023learning}.
We empirically found this benefits the geometry quality. The  mesh-anchored hash tables $\hat{\hashTable}_{ij}$ and features $\vertFeat_{ij}$ are decoded into a neural radiance field as described in \cref{sec:nerf_dec}.

\begin{figure}[htp!]
    \centering
    \includegraphics[width=0.95\linewidth]{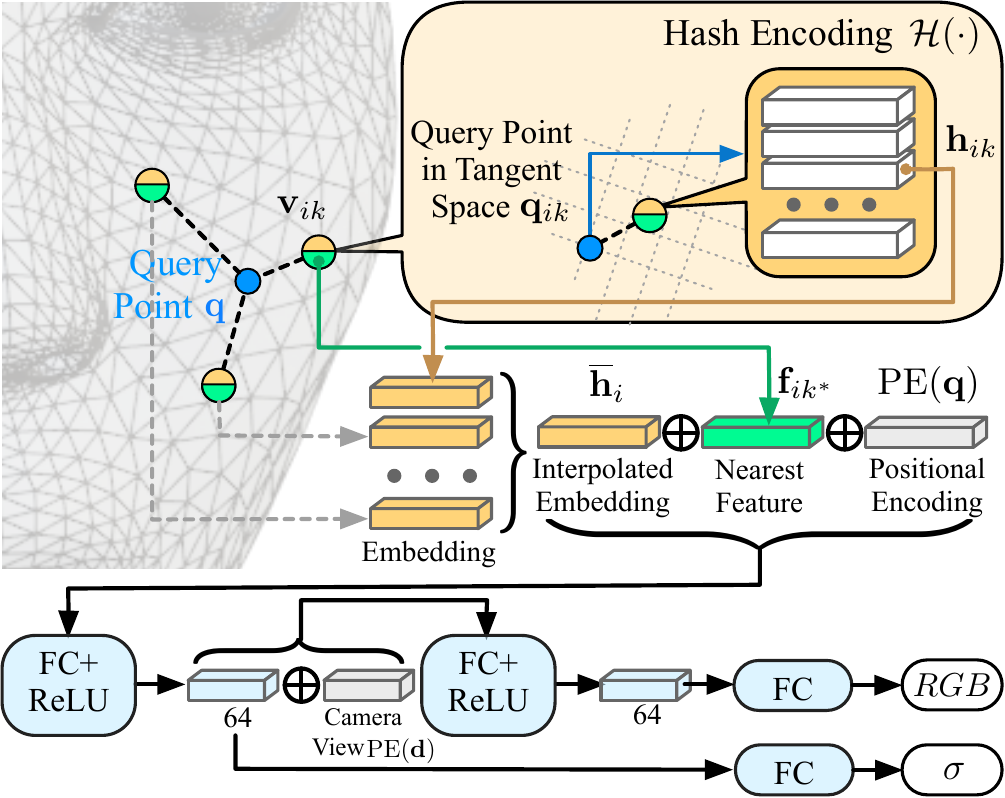}
    \caption{Pipeline of Neural Radiance Field Decoding (\cref{sec:nerf_dec}). Our method only needs a very lightweight MLP for NeRF inference, where a two-hidden-layer MLP is used to predict the density and the color of each query point.}
    \label{fig:nerf_dec}
\end{figure}

\subsection{Neural Radiance Field Decoding}
\label{sec:nerf_dec}
Given the mesh-anchored hash tables $\hat{\hashTable}_{ij}$ and features $\vertFeat_{ij}$ described in \cref{sec:merge_blendshape}, the final step is to decode them into a Neural Radiance Field (NeRF) to render the output image as shown in \cref{fig:nerf_dec}. The key idea is to associate a query point to neighbor vertices, and pull the embeddings from the attached hash tables via hash encoding~\cite{mueller2022instant}. Finally, we decode these pulled embeddings and the nearest per-vertex feature into the color and the density of this query point, followed by volumetric rendering to obtain the output image.

For a 3D query point $\queryPoint$ when rendering a particular facial expression $i$, we first obtain its $k$-nearest-neighbors, denoted as $\{\verts_{i\jKNN}\}_{\jKNN \in \mathcal{N}_{\queryPoint}^K}$, from the 3DMM vertices, with $\jKNN^{\ast}$ denoting the nearest vertex index.
For each neighbor vertex $\verts_{i\jKNN}$ with an attached hash table $\hat{\hashTable}_{i\jKNN}$, we denote $\queryPoint_{i\jKNN}$ as the coordinates of $\queryPoint$ in the tangent space of $\verts_{i\jKNN}$.
We then use $\queryPoint_{i\jKNN}$ to query the hash table $\hat{\hashTable}_{i\jKNN}$ using a hash encoding function $\mathcal{H}(\cdot)$ and obtain the embedding $\hashTableEmbedding_{i\jKNN} = \mathcal{H}(\queryPoint_{i\jKNN}; \hat{\hashTable}_{i\jKNN})$.
To interpolate the embeddings from all $k$-nearest-neighbors, we use the weighted sum of the inverse distances $z_\jKNN = 1 / \| \queryPoint_{i\jKNN} \|_2$. 
Next, the summed embedding, together with the nearest per-vertex feature $\vertFeat_{i\jKNN^{\ast}}$ and the query point tangent coordinate $\queryPoint_{i\jKNN^{\ast}}$ of the nearest vertex, are fed into a two-hidden-layer MLP to predict the density and color as
\begin{gather}
    \overline{\hashTableEmbedding}_{i} = \sum_{\jKNN \in \mathcal{N}_{\queryPoint}^K} \overline{w}_{\jKNN} \hashTableEmbedding_{i\jKNN}, \mathrm{ where }~\overline{w}_{\jKNN} = \frac{z_\jKNN}{\sum_{{\jKNN^\prime} \in \mathcal{N}_{\queryPoint}^K} z_{\jKNN^\prime}},\\
    \Bigl[\mathbf{c}_i (\queryPoint, \cameraDir), \sigma_i (\queryPoint)\Bigr] = \mathcal{F}_{\mathrm{MLP}} \left(\overline{\hashTableEmbedding}_i, \vertFeat_{i\jKNN^{\ast}}, \mathrm{PE}(\queryPoint_{i\jKNN^{\ast}}), \mathrm{PE}(\cameraDir) \right), \nonumber
\end{gather}
where $\overline{w}_\jKNN$ is the normalized inverse-distance based weight, $\cameraDir$ denotes the camera view direction, $\mathrm{PE}(\cdot)$ denotes positional encoding, $\mathbf{c}_i$ denotes color, and $\sigma_i$ denotes density. Finally, we render the output pixel with the given camera ray $\cameraRay$ by volumetric rendering, where we reparameterize the query point with samples on the ray $\queryPoint = \cameraRay(t) = \mathbf{o} + t \cameraDir$:
\vspace{-0.8em}
\begin{align}
    \mathbf{C}_i(\cameraRay) &= \int_{t_n}^{t_f} T(t) \sigma_i(\cameraRay(t)) \mathbf{c}_i(\cameraRay(t), \cameraDir)  dt, \nonumber \\
    \mathrm{where}~T(t) &= \mathrm{exp} \left( - \int_{t_n}^{t} \sigma_i(\cameraRay(s)) ds \right).
\end{align}
Following prior works \cite{jiang2022neuman,bai2023learning}, we also introduce a per-frame error-correction warping field during training to reduce misalignments due to the noise in 3DMM tracking and unmodeled per-frame contents such as hair movements. We feed the query point $\queryPoint$, together with a per-frame latent code $\mathbf{e}_i$, into an MLP $\mathcal{F}_{\mathcal{E}}(\cdot)$ to obtain a rigid transformation applied on the original query point, denoted as $\queryPoint^{\prime} = \mathcal{T}_i(\queryPoint) = \mathcal{F}_{\mathcal{E}}(\queryPoint, \mathbf{e}_i)$. The warped query point $\queryPoint^{\prime}$ is then used to compute the density and color for volumetric rendering. Since the warping fields are overfit to corresponding training frames, we disable the warping field during testing similar to previous works \cite{jiang2022neuman,bai2023learning}, and hence $\mathcal{F}_{\mathcal{E}}$ does not affect rendering efficiency.

\subsection{Hierarchical $k$-nearest-neighbor Search}
\label{sec:hiera_knn}
As described in \cref{sec:nerf_dec}, our method involves a $k$-nearest-neighbor ($k$-NN) search, which is computationally expensive and cannot be naively accelerated with pre-calculated structures (\eg, KD-Tree) due to the dynamically changing search pool (\ie, the 3DMM vertices driven by poses and expressions). 
To speed up the process, we propose a hierarchical $k$-NN search algorithm following a coarse-to-fine strategy. 
The key idea is to group nearby query points into a cluster as they likely share similar nearest neighbors.
% Then we search for $K^\prime$ nearest neighbors (where $K^\prime>K$) for each cluster, from which each query point within the cluster searches for $K$ nearest neighbors.
% The key idea is to define a 3D voxel grid, and group neighbor query points into the voxel they fall in, then run $K$-NN search for the voxel center to obtain $K$ candidate nearest neighbor vertices (where $K>k$) from a total of $N$ vertices in the 3DMM. Then, for each query point, we run $k$-NN search from the $K$ candidates of its voxel instead of the whole $N$ vertices.
Specifically, we use a 3D grid with resolution 64 and treat all query points that fall in each voxel as a cluster. 
%Each cluster searches for $K=12$ nearest neighbors using the voxel center from $N=2047$ 3DMM vertices, and then search $k=3$ neighbors for each query point. 
For each cluster, we first search $K^\prime$  (where $K^\prime>K$) nearest neighbors of the voxel center from all 3DMM vertices.
Then, for each query point, we search $K$ nearest neighbors from the $K^\prime$ nearest neighbors of the corresponding cluster.
In practice, for a 3DMM with the vertices number of $J=1772$, we set $K^\prime=12$ and $K=3$.
Our experiments empirically show that, with a proper grid resolution, this design significantly improves the nearest neighbor search speed, and does not introduce noticeable rendering artifacts, even though the $k$-NNs may not be accurate on some of the query points.
% necessary to further accelerate our appoarch to real-time rendering (\ie, average $\ge 30$ FPS to render a video).
% Even though this may produce inaccurate $k$-NNs on some query points, we empirically find this does not introduce any noticeable rendering artifacts when the 3D grid resolution is large enough (\ie, $64$ as we used here).

\subsection{Training}
\label{sec:train}
Only monocular RGB videos are required to train our model. Three losses are used during the training process: (1) a photometric loss that minimizes the $l_2$-norm distance between the rendered and ground truth pixel colors over all camera rays $\mathbf{r}$ from all training frames $i$. Formally, we have $\mathcal{L}_\mathrm{rgb} = \sum_i \sum_{\mathbf{r}} \| \mathbf{C}_i(\mathbf{r}) - \mathbf{I}_i(\mathbf{r}) \|_2$; (2) a elastic regularization loss $\mathcal{L}_\mathrm{elastic}$ applied on the learned error-correction warping field $\mathcal{T}(\queryPoint)$, which is introduced in Nerfies~\cite{park2021nerfies}; (3) a magnitude regularization loss to encourage small warping fields, which is defined as $\mathcal{L}_\mathrm{mag} = \sum_{\queryPoint} \| \queryPoint - \mathcal{T}(\queryPoint) \|_2^2$. Finally, we combine all three loss terms:
\begin{align}
    \mathcal{L} = \mathcal{L}_\mathrm{rgb} + \lambda_\mathrm{elastic} \mathcal{L}_\mathrm{elastic} + \lambda_\mathrm{mag} \mathcal{L}_\mathrm{mag},
\end{align}
where we set $\lambda_\mathrm{elastic} = 10^{-4}$ at the beginning of the training and decay it to $10^{-5}$ after $150$k iterations, and set $\lambda_\mathrm{mag} = 10^{-2}$. To warm start training, we replace the $l_2$-norm distance in the photometric loss $\mathcal{L}_\mathrm{rgb}$ with the $l_2$ distance for the first $10$k iterations. Please refer to the supplementary for more details.

% \subsection{Reducing Query Sampling via Surface Bonded Boxes}

\if
get a controllable version of NGP: By deforming the FLAME mesh with pose and expression codes, 

which is a online controllable version of NGP conditioned on the FLAME poses and expressions.

the rendering quality of Parametric Model Driven Avatar~\cite{ourcvpr2023}, and the rendering speed of NGP~\cite{}.

The core module to achieve this goal is the Expression Dependent Local NGP (\cref{sec:exp_dep_local_ngp}).

May move to sec3.2

The idea is to attach a local 3D feature grid to each vertex of the FLAME mesh, and efficiently represent the 3D feature grid with multi-level hash tables as in NGP~\cite{mueller2022instant}. For convenience, we denote these "hash table represented grids" as "NGP grids". As a result, we obtain a controllable version of NGP: By deforming the FLAME mesh with pose and expression codes, these NGP grids will be moved together with their associated vertices, making the decoded radiance field controllable by the given pose and expression codes. However, merely moving the NGP grid with FLAME motions can only model the coarse level deformations of poses and expressions but fails to capture fine-grained details such as wrinkles, since the detail contents inside each NGP grids are fixed. To this end, we further condition the NGP grids on poses and expressions to better model detailed pose and expression variations, by 
\fi

\if
To combine the advantages from both preliminaries (\cref{sec:prelim}) (\ie, high rendering quality, controllability, and fast rendering speed), we propose the Expression Dependent Local NGP (\cref{sec:exp_dep_local_ngp}). The core idea is to locally attach a small NGP hash table on each FLAME vertex to encode the neural radiance field anchored on the vertex. These NGP encoded NeRFs can thus be deformed according to the motion of the FLAME mesh driven by the pose and expression codes, inheriting the controllability from Parametric Model Driven Avatar~\cite{ourcvpr2023}. To capture details (\eg, expression-dependent wrinkles) that cannot be modeled by the coarse FLAME mesh motions, we further condition the hash table features on the FLAME poses and expressions, by attaching multiple NGP hash tables to each vertex and linearly blending them with weights predicted from CNN in UV space. The CNN inherited from Bai \etal \cite{ourcvpr2023} brings in the high rendering quality, while the NGP encoded NeRFs enables fast rendering speed.
\fi

\if
\subsection{Preliminary}
\label{sec:prelim}

\subsubsection{Parametric Model Driven Avatar}
To better introduce our method, we leverage the Parametric Model Driven Avatar proposed by Bai \etal \cite{ourcvpr2023} as a starting point, which is a 3DMM-anchored neural radiance field controlled via the FLAME poses and expressions, and adapt two main components of their pipeline: \textit{Predicting expression-dependent local features}, and \textit{Neural Radiance Field Decoding}.

\paragraph{Predicting expression-dependent local features}. Given a specified FLAME pose codes $\bm{\theta}_i$ and expression codes $\bm{\psi}_i$, the deformed FLAME mesh is computed as $\bm{V}_t=(\bm{\psi}_t, \bm{\theta}_t)$, where $i$ denotes the frame index and the shape code is ignored for brevity. To obtain the expression-dependent local features, we first compute the FLAME vertex displacements $\bm{D}_i$ between the deformed mesh $\bm{V}_t$ and the neural mesh $\bm{V}_{neutral}$ as $\bm{D}_i = \bm{V}_i(\bm{\psi}_i, \bm{\theta}_i) - \bm{V}_{neutral}(\bm{0}, \bm{0})$. The displacements are then rasterized into UV space and feed into a U-Net to obtain a feature map. The feature map is then sampled back to FLAME vertices, serving as the expression-dependent local feature $\{\bm{z}_i^j\}$ attached on each FLAME vertex with $j$ denoting the vertex index.

\paragraph{Neural Radiance Field Decoding}. Given the deformed FLAME mesh $\bm{V}_t$ with attached local features $\{\bm{z}_i^j\}$, we can decode them into a radiance field and obtain the output image with volumetric rendering. For a $3$D query point,

\subsubsection{NGP}
\fi

\section{Experiments}

In this section, we first introduce the data and metrics used for training and evaluation (\cref{sec:data_metrics}). 
Then, we show that our avatar model achieves real-time rendering speed, while producing superior rendering quality on challenging expressions than recent efficient avatars~\cite{Gao2022nerfblendshape,zielonka2023instant,zheng2023pointavatar} and being comparable to previous high-quality approaches~\cite{bai2023learning} (\cref{sec:compare_sota}).
% Then, we show that our avatar model produces a superior rendering quality than recent efficient avatars~\cite{Gao2022nerfblendshape,zielonka2023instant} and is comparable to previous high-quality approach~\cite{bai2023learning}, while achieving a real-time rendering speed (\cref{sec:compare_sota}).
Finally, we provide ablation studies to justify the design choices and hyper-parameters of our avatar representation, and demonstrate the rendering speed improvements contributed from each of our newly proposed algorithmic components.
% We then show that our model is able to deliver premium rendering quality with comparable computation costs than existing works.

\subsection{Data and Metrics}
\label{sec:data_metrics}
\paragraph{Data.}
We use monocular RGB videos of multiple subjects (\ie, one video for one subject) to train and evaluate our method, and compare to prior state-of-the-art (SOTA) approaches. 
% We adopt $5$ subjects from MonoAvatar~\cite{bai2023learning}, $1$ subject from PointAvatar~\cite{zheng2023pointavatar}, $1$ subject from INSTA~\cite{zielonka2023instant}, and we additionally capture videos of $3$ subjects following the same protocol as MonoAvatar~\cite{bai2023learning}. 
Our dataset consists of $10$ videos in total, which are a mix of videos captured by us, as well as videos from prior works including PointAvatar~\cite{zheng2023pointavatar}, INSTA~\cite{zielonka2023instant}, and MonoAvatar~\cite{bai2023learning}.
We filter out the background of the videos with off-the-shelf segmentation~\cite{lugaresi2019mediapipe} and matting~\cite{lin2022robust} methods, then crop and resize the videos into a VGA resolution that preserves the original aspect ratio. We compute the camera and 3DMM parameters from the videos following the 3DMM fitting optimization used in INSTA~\cite{zielonka2023instant}. We reserve a short clip from the end of each video as the testing frames, and use the rest frames for training.

% For fairly comparisons with prior arts, we use the same dataset and preprocessing procedures as presented in Anonymous \etal~\cite{ourcvpr2023} to train and evaluate our model. 
% The dataset contains monocular RGB videos captured by the authors of Anonymous \etal~\cite{ourcvpr2023} from $4$ subjects with vivid and challenging facial expressions and head pose, and a $2$ minutes talking head video with 50fps from NerFACE~\cite{gafni2021dynamic} which is in talk head fashion and has relatively simple expressions.
% We obtain the camera and 3DMM parameters with off-the-shelf $3$DMM fitting optimization and filter out the background and adopt the same training and testing frame split following Anonymous \etal~\cite{ourcvpr2023}.

\paragraph{Metrics.}
Following prior arts~\cite{gafni2021dynamic}, we use PSNR, SSIM (higher is preferable), and LPIPS (lower is preferable) to measure the image quality.
As observed by Zhang \etal~\cite{zhang2018unreasonable}, LPIPS is a more effective metric in judging the perceptual quality compared to PSNR and SSIM.
When computing PSNR and SSIM, we weigh the mean squared error map and the SSIM map with a foreground mask (eroded and smoothed), in order to focus on non-empty areas and avoid the inaccurate foreground segmentation from dominating the metrics.

To evaluate the computational cost, we measure the rendering speed in frames-per-second (FPS) on a RTX3090Ti and compare across different approaches with their available implementations.
We also estimate the number of FLOPs (floating-point operations) of all methods as the theoretical measurement for the rendering speed. When estimating FLOPs, we fix the contribution of the ray-marching part to $16$ points sampled along each camera ray. This simplifies the estimate, since the ray-marching varies the number of FLOPs needed across cameras and scenes, and it applies to all the considered methods.

\begin{figure*}[htp!]
    \centering
    \includegraphics[width=\textwidth]{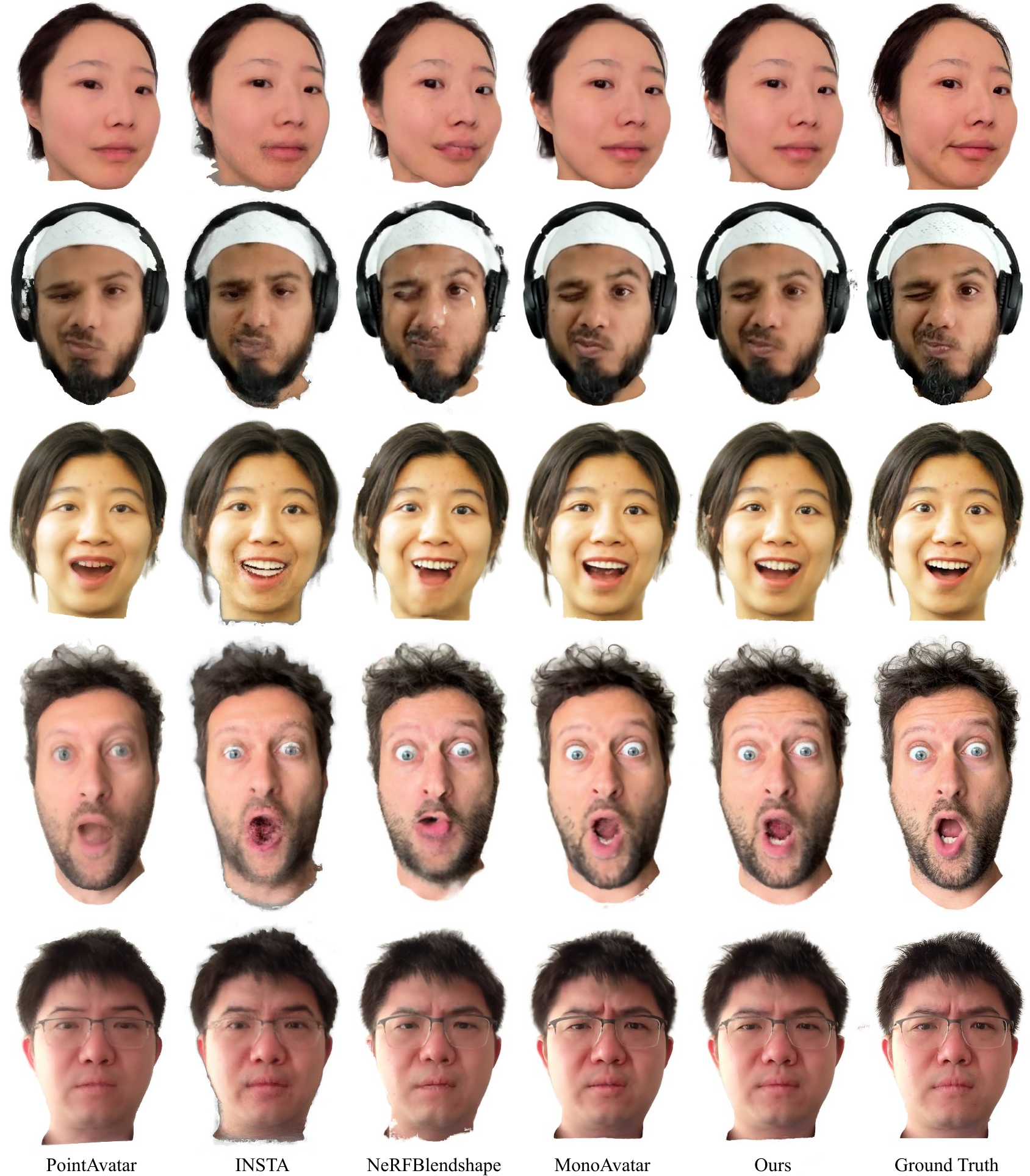}
    \caption{Comparisons on the rendering quality to previous state-of-the-art methods. From left to right, each column contains the images of: 1) PointAvatar~\cite{zheng2023pointavatar}, 2) INSTA~\cite{zielonka2023instant}, 3) NeRFBlendshape~\cite{Gao2022nerfblendshape}, 4) MonoAvatar~\cite{bai2023learning}, 5) Ours, 6) Ground Truth. Our method faithfully reconstructs the personalized expressions and high-frequency details, achieving one of the best rendering quality with real-time rendering speed.}
    \label{fig:comparison_full}
\end{figure*}

\begin{figure*}[htp!]
% \vspace{-3mm}
    \centering
    \includegraphics[width=\textwidth]{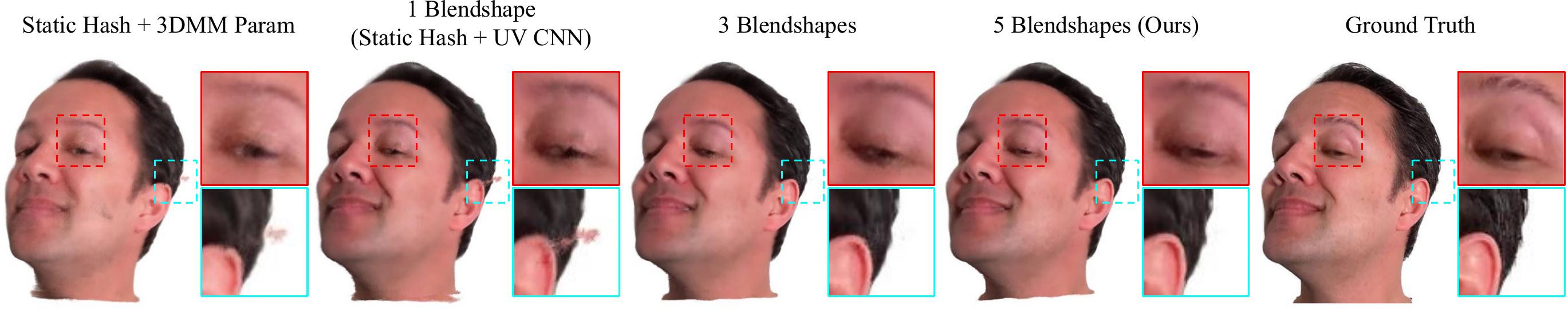}
    \vspace{-2em}
    \caption{Qualitative results of our full model and various alternative design choices to demonstrate the necessity of our mesh-anchored hash table blendshapes. Please refer to \cref{sec:ablation_hash_blendshape} for details of each alternative design choice. ``\textit{Static Hash + 3DMM Param}'' gives unfaithful expressions and obvious artifacts around eyes and ears. ``\textit{Static Hash + UV CNN}'' produces more faithful expressions, but still suffers from floaters. By utilizing the blendshape formulation of hash tables, ``\textit{3 Blendshapes}" eliminates most of the artifacts, but produces blurriness on the details of eyes and ears. ``\textit{Ours}" gives the best rendering quality by increasing the number of blendshapes to $5$, leading to cleaner details of the eyelids and ear boundary.}
    \label{fig:ablations}
    \vspace{-4pt}
\end{figure*}

\subsection{Comparison to State-of-the-art}
\label{sec:compare_sota}

We compare our method with several prior works, including: NeRFBlendshape~\cite{gafni2021dynamic}, INSTA~\cite{zielonka2023instant}, PointAvatar~\cite{zheng2023pointavatar}, and MonoAvatar~\cite{bai2023learning}.
NeRFBlendshape~\cite{gafni2021dynamic} and INSTA~\cite{zielonka2023instant} adopt hash encoding~\cite{mueller2022instant} into head avatars, leading to efficient renderings. PointAvatar~\cite{zheng2023pointavatar} leverages point clouds to represent the head avatar. MonoAvatar~\cite{bai2023learning} is based on a 3DMM-anchored NeRF, and produces high-quality renderings but is slow in speed. We use the same camera and 3DMM parameters to train and test all methods.

From \cref{tab:sota}, ours and INSTA~\cite{zielonka2023instant} are the only 2 methods that can achieve real-time rendering (\ie, $\ge 30$ mean FPS). However, INSTA is quantitatively inferior than our method by a large margin, and gives obvious artifacts in \cref{fig:comparison_full}, especially for challenging expressions. PointAvatar~\cite{zheng2023pointavatar} has the potential to run in real-time with an optimized implementation thanks to its point cloud representation, but their renderings are overall blurrier than ours, leading to worse quantitative results. Although NeRFBlendshape~\cite{Gao2022nerfblendshape} gives relatively good numbers in \cref{tab:sota}, it produces severe artifacts in dynamically changing regions (\eg, mouth and eyebrows in \cref{fig:comparison_full}) for several median and large expressions and also gives more floaters, resulting in implausible animations. We highly suggest readers to see the supplementary videos for more comparisons. 
% and qualitatively evaluate temporal stability of the approaches. 
MonoAvatar~\cite{bai2023learning} gives good rendering qualities and animations, but is one order of magnitudes slower and slightly blurrier on high frequency details such as forehead wrinkles, presumably because that the hash encoding in our model can better capture high frequency contents. Among these compared approaches, our method is the only one that achieves real-time rendering while being one of the best on image qualities.

We also compare the theoretical FLOPs of all methods in \cref{tab:sota}, where our method requires the least computation mostly because of the smaller MLPs we use (\eg, $2$ hidden layers of ours vs. $\ge 5$ hidden layers of others). Note that INSTA~\cite{zielonka2023instant} is implemented with a highly optimized pure C++ and CUDA codebase, while other methods use python (tensorflow/pytorch) with customized CUDA kernels. This implementation advantage of INSTA makes it running in a high FPS even with a relatively larger FLOPs.

\begin{figure*}[h!]
% \vspace{-3mm}
    \centering
    \includegraphics[width=\textwidth]{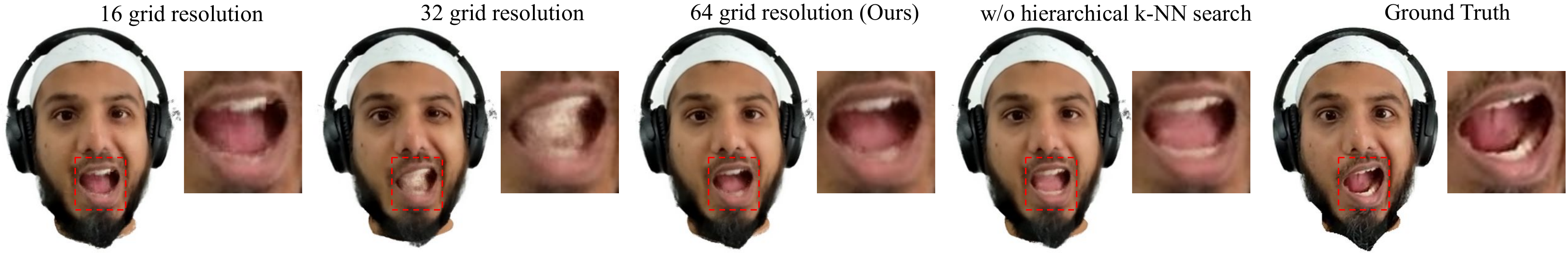}
    \vspace{-2em}
    \caption{Qualitative results of using different 3D grid resolutions during hierarchical $k$-NN search and without hierarchical $k$-NN search for ablation studies. ``\textit{16 grid resolution}" results in blocky artifacts around mouth and nose. ``\textit{32 grid resolution}" removes blocky artifacts, but produces floaters inside mouth. ``\textit{64 grid resolution}'' produces a similar level of visual quality as ``\textit{w/o hierarchical $k$-NN search}''.}
    \label{fig:ablation_knn}
    \vspace{-10pt}
\end{figure*}

\begin{table}[t]
\centering
\small
\setlength{\tabcolsep}{0.3em}
\begin{tabular}{l|ccc|cc}
\toprule
 & {\footnotesize LPIPS} & {\footnotesize SSIM} & {\footnotesize PSNR} & {\footnotesize Mean FPS} & {\footnotesize GFLOPs} \\
\midrule
% \multicolumn{5}{c}{\footnotesize \textit{Full Training Data}} \\
% \hline

PointAvatar~\cite{zheng2023pointavatar}    & 0.117 & 0.728 & 21.12 & 5.0 & 933 \\

INSTA~\cite{zielonka2023instant}   & 0.149 & 0.758 & 22.12 & \textcolor{Green}{\textbf{46.2}} & 266 \\

NeRFBlendshape~\cite{Gao2022nerfblendshape}    & 0.110 & 0.793 & \textbf{22.77} & 11.2 & 223 \\

MonoAvatar~\cite{bai2023learning}    & 0.114 & \textbf{0.798} & 22.74 & 0.5 & 2385 \\

Ours    & \textbf{0.100} & 0.795 & \textbf{22.77} & \textcolor{Green}{\textbf{35.9}} & \textbf{113} \\

\bottomrule

\end{tabular}
\vspace{-1em}
\caption{Quantitative comparisons with state-of-the-art approaches. Our method achieves rendering quality among the best, while supports real-time rendering with a $512 \times 512$ resolution.}
\label{tab:sota}
\vspace{-1em}
\end{table}

\subsection{Ablation Study}
In this section we show the impact of the proposed design choices, in particular proving the importance of mesh-anchored hash table blendshapes and the proposed hierarchical $k$-NN search.

\subsubsection{Mesh-anchored Hash Table Blendshapes}
\label{sec:ablation_hash_blendshape}
We hereby investigate alternative design choices and different hyper-parameters to justify the necessity of our mesh-anchored hash table blendshapes.

% \vspace{0.5mm}
\noindent\textbf{Static Hash + 3DMM Param.}
We first build a naive alternative approach to incorporate hash encoding into 3DMM-anchored NeRF, by attaching a single hash table to each vertex. For a query point, we concatenate the 3DMM pose and expression codes ($\bm{\theta}_i, \bm{\psi}_i$) with the embedding pulled from the hash tables, and send them into the MLP.
Note that there is no convolution running in UV space.
As shown in \cref{tab:ablation} and \cref{fig:ablations}, obvious rendering artifacts show up, and the rendering quality metrics drop significantly compared to our full model.
This is presumably because the lightweight MLP, which is crucial for good efficiency, does not have enough capacity to process detailed expression-dependent information from compact 3DMM codes.

% \zb{Only coarse draft after here.}

% \vspace{0.5mm}
\noindent\textbf{Static Hash + UV CNN.}
We then increase the model capacity by adding back the UV CNN branch, but use only one hash table per-vertex, \ie, single hash table without blendshape formulation.
As shown in \cref{tab:ablation} and \cref{fig:ablations}, the rendered images show less artifacts over the previous case, but still contain blurry textures and floaters compared to our full model.
This demonstrates that the blendshape formulation of mesh-anchored hash tables are necessary in order to obtain good expression dependent local embeddings, leading to a superior rendering quality.

% \vspace{0.5mm}
\noindent\textbf{Number of Hash Table Blendshapes.}
Here we investigate how the number of hash table blendshapes per vertex influences the final rendering quality. As shown in \cref{tab:ablation}, we can see that more blendshapes per vertex leads to higher rendering qualities, which saturates as the number of tables increases. In \cref{fig:ablations}, increasing the number of blendshapes also gives better details especially for eyelids and ears. Although further adding more blendshapes may produce a better quality, we choose to use $5$ blendshapes per-vertex as our final setting to maintain a relatively small model size and computation cost.

\subsubsection{Hierarchical $k$-NN Search}
\label{sec:ablation_hiera_knn}
We evaluate our model with and without hierarchical $k$-NN search in terms of speed and quality. From the comparison over rendering speeds (w/: $35.9$ FPS; w/o: $26.4$ FPS),
we can see that hierarchical $k$-NN search gives around $36\%$ improvements on the frame rate, which is crucial for achieving real-time rendering.
From \cref{fig:ablation_knn}, we empirically find that enabling hierarchical $k$-NN search will not lead to observable drops on the rendering quality, as long as a proper grid resolution is used (\ie, $64$ in our case).

We also investigate the affect of using different 3D grid resolutions during hierarchical $k$-NN search. As shown in \cref{fig:ablation_knn}, we observe more artifacts around the mouth region when using smaller 3D grid resolutions (\ie, $32$ and $16$). Therefore, we choose to use $64$ resolution in our final setting, which is a good trade-off between quality and speed.

\begin{table}[t]
\centering
\small
\setlength{\tabcolsep}{0.35em}
\begin{tabular}{l|c|c|c}
\toprule
 & LPIPS & SSIM & PSNR \\
\midrule
% \multicolumn{5}{c}{\footnotesize \textit{Full Training Data}} \\
% \hline

Static Hash + 3DMM Param    & 0.125 & 0.763 & 21.99 \\

1 Blendshape (Static Hash + UV CNN)   & 0.115 & 0.785 & 22.52 \\

3 Blendshapes    & 0.104 & 0.791 & 22.70 \\

5 Blendshapes (Ours)    & \textbf{0.100} & \textbf{0.795} & \textbf{22.77} \\

\bottomrule

\end{tabular}
\vspace{-0.75em}
\caption{Quantitative results for the ablation study. Our full model (\ie, mesh-anchored hash table blendshapes) consistently outperforms alternative design choices discussed in \cref{sec:ablation_hash_blendshape}.}
\label{tab:ablation}
\vspace{-1.25em}
\end{table}

\section{Discussion}
We present a high quality 3D neural volumetric head avatar that can be rendered efficiently, while only requires monocular RGB videos for construction.
% in real-time at $512 \times 512$ resolution.
We propose the mesh-anchored hash table blendshapes as our avatar representation, which enable a significantly faster rendering speed by utilizing hash encoding and lightweight MLPs, while still maintaining superior controllability to support realistic facial animations, and producing vivid expression-dependent details thank to the local blendshape formulation of hash tables.
The experiments indicate that our approach runs in real-time at a $512 \times 512$ resolution, while giving a rendering quality comparable to state-of-the-art, with better challenging expressions than prior efficient approaches.
% We further show that an expression driven NGP, achieved by learning multiple local instant NGP combined via weights predicted in the UV space of a 3DMM, is effective in delivering expression-dependent features for fine rendering details. To further speed up rendering, a hierarchical nearest neighbor search approach is proposed as an approximate solution, but sufficient to support avatar rendering without loss in quality.

As a limitation, we observe floaters under camera viewpoints and expressions that are far from the training distribution, which is a common issue in instant NGPs~\cite{mueller2022instant} based approaches.
% Adding more regularization and learning objectives may potentially mitigate the problem.
We also notice that performance around the mouth interior regions tends to be less stable because of the relatively poor tracking in these areas on the training data.
Fortunately, the fast rendering could enable the possibility to adopt more expensive training strategies, such as regularization terms, adversarial loss, or joint face fitting refinement during the training, which could potentially mitigate these issues and further improve the rendering expressiveness and quality.

{
    \small
    \bibliographystyle{ieeenat_fullname}
    \bibliography{main}
}

% Supp
\clearpage

\begin{center}
\vspace{10mm}
\textbf{\Large Supplementary Materials}
\end{center}
\renewcommand\thesection{\Alph{section}}
\renewcommand\thetable{\Alph{table}}
\renewcommand\thefigure{\Alph{figure}}

\setcounter{section}{0}
\setcounter{figure}{0}
\setcounter{table}{0}

In this supplementary material, we provide additional method details and more results, including hash encoding details (\cref{sec:hash_enc}), warping fields details (\cref{sec:warp}), network architectures (\cref{sec:arch}), training and testing details (\cref{sec:train_details}), data statistics (\cref{sec:data}), additional experiments (\cref{sec:results} and the supplementary webpage), as well as discussions on limitations (\cref{sec:limit}).
% webpage: https://drive.google.com/file/d/1joG-PBNpfH7PjziW-k-ulSbT1_rPXFNo/view?usp=drive_link&resourcekey=0-QwYeK5CxX7GapiAt2sERMA

\section{Hash Encoding Details}
\label{sec:hash_enc}

As described in Sec.3.1 and 3.2, we attach local hash table blendshapes on each 3DMM vertex, which are linearly blended with expression-dependent weights predicted via the U-Net running in UV space as the merged hash table for each vertex. The hyper-parameters of hash tables are shown in \cref{tab:hashtab}.

% \begin{table*}[!t]

% \small
% \setlength{\tabcolsep}{0.6em}
% \begin{tabular}{l|c|c|c|c|c}
% \toprule
%     & NeRFBlendshape~\cite{} & Anonymous \etal~\cite{}  & Anonymous \etal~\cite{} w/ Fast kNN & Ours w/o Fast kNN & Ours \\

% \midrule

% Constant FLOPS(G) per-frame    & 9.8566\times10^{-2} & - & - & 9.8744 & 9.8744 \\

% FLOPS(G) per-query point   & 5.7664\times10^{-5} & - & - & 5.3625\times10^{-5} & 3.4590\times10^{-5} \\

% \midrule

% \makecell{Total FLOPS(G) with \\ $512$ * $512$ * $64$ \\ query points per-frame}   & 967.8252 & - & - & 909.795220672 & 590.4898 \\
% \bottomrule
% \end{tabular}
% \caption{Computation costs comparisons to various baselines, in terms of the required FLOPS (floating-point operations) of a method.}
% \label{tab:sota_flop}
% \end{table*}

\begin{table}[!h]
% \small
\centering
\setlength{\tabcolsep}{0.8em}
\begin{tabular}{l|c}
\toprule
    Parameters & Values \\
\midrule
    Number of levels & $2$ \\
    Hash table size & $2^8$ \\
    Number of feature channels & $4$ \\
    Coarsest resolution & $32$ \\
    Finest resolution & $64$ \\
    Number of blendshapes per-vertex & $5$ \\
\bottomrule
\end{tabular}
% \vspace{1mm}
\caption{Hyper-parameters for mesh-anchored hash table blendshapes.
}
\label{tab:hashtab}
\end{table}

Instead of attaching hash tables to all 3DMM vertices (\ie FLAME~\cite{FLAME:SiggraphAsia2017} in our work), we select a subset of vertices to reduce the computation and model size, as well as ensure a more uniform vertex distribution on the 3DMM surface. More specifically, we subsample the vertices using poisson-disk sampling from meshlab~\cite{meshlab} on a template mesh without eyeballs, and manually add 10 iris vertices, resulting in $1772$ vertices in total.

\section{Warping Field Details}
\label{sec:warp}

As described at the end of Sec.3.3, following prior works on deformable NeRF~\cite{jiang2022neuman,weng2022humannerf,bai2023learning}, we overfit 3D warping fields on training frames to alleviate the negative influence of misalignments between tracked 3DMM meshes and images due to the noise in 3DMM tracking and unmodeled per-frame contents such as hair movements. These warping fields are discarded during testing as in \cite{jiang2022neuman} and \cite{bai2023learning} since they are overfit to training frames.

More specifically, we first assign a learnable latent code $\mathbf{e}_i$ for each training frame $i$. Given a query point $\mathbf{q}$, we apply positional encoding on its coordinates and concatenate with the latent code $\mathbf{e}_i$, then feed them into an MLP $\mathcal{F}_{\mathcal{E}}(\cdot)$ to obtain a rigid transformation consists of 3 components: a $3$D rotation $\mathbf{R} \in SO(3)$ , a rotation center $\mathbf{c}^{rot}$, and a $3$D translation $\mathbf{t}$. We then compute the warped query point $\mathbf{q}^{\prime}$ by applying the rigid transformation to the original query point as
\begin{align}
    \mathbf{R}, \mathbf{c}^{rot}, \mathbf{t} &= \mathcal{F}_{\mathcal{E}} \left( \mathbf{q}, \mathbf{e}_i \right) \\
    \mathbf{q}^{\prime} &= \mathbf{R} \left( \mathbf{q} + \mathbf{c}^{rot} \right) - \mathbf{c}^{rot} + \mathbf{t}.
\end{align}
In practice, we represent $\mathbf{R}$ with a pure log-quaternion and directly regress it with the MLP $\mathcal{F}_{\mathcal{E}}(\cdot)$. As described in Sec.3.3, we denote this full warping procedure as $\mathbf{q}^{\prime} = \mathcal{T}_i(\mathbf{q}) = \mathcal{F}_{\mathcal{E}}(\mathbf{q}, \mathbf{e}_i)$. The warped query point $\mathbf{q}^{\prime}$ is then used to compute the density and color for volumetric rendering.

\section{Network Architectures}
\label{sec:arch}

% The details of the proposed architecture are shown in Figure . 
Here we introduce the detailed network architectures of three main components described in Sec.3.2 and 3.3: the U-Net running in UV space, the MLP to predict densities and colors, and a warp field MLP predictor. Please refer to the main paper for details on how these components come together to form our avatar model.

\subsection{U-Net running in UV space}
\label{sec:unet}
As described in Sec.3.2, the U-Net running in UV space takes the 3DMM vertex displacements as the input and outputs expression-dependent weights (to weighted sum hash tables) and per-vertex features. For the encoder side, we use downsampling residual blocks to extract a feature pyramid with the number of channels for each level as $\{8, 16, 32, 64, 128, 256 \}$, with 128 as the input resolution and downsample to 64, 32, 16, 8, 4, 2. In the decoder side, we use upsampling residual blocks (\ie with transposed convolutions) and set the number of output channels for each level to 128, 64, 64, 64, 64, 64. Finally, we use a $1 \times 1$ convolution layer to get the weights map and the feature map. The leaky ReLU is applied after each convolutional layer with a slope $0.2$. The input vertex displacement map has a resolution of $128 \times 128$. The output expression-dependent weights map has a size of $128\times128\times5$ ($4$ channels predicted by network, $1$ channel set to a constant one) and the output feature map has a size of $128\times128\times24$. The weights map and the feature map are then sampled back to 3DMM vertices as described in Sec.3.2.

% Our expression-dependent weight predictor is a 6-level residual U-Net. We use residual blocks to extract feature, and the feature channels of each level are set as 4, 8, 16, 32, 64, 128. In the decoder, residual blocks with transposed convolutions are applied to increase the spatial resolution. The leaky ReLU is applied after each convolutional layer with slope $0.2$. The input of the predictor is a 3D deformation map in $128 \times 128$ resolution which stores the vertex displacement from the neural expression to the current facial expression in UV space, and the output of the U-Net are a weight map $128\times128\times5$ and a feature map $128\times128\times16$.

\subsection{MLP for Densities and Colors}
As described in Sec.3.3, for each query point, we use the tiny MLP to decode densities and colors from the summed hash table embedding $\overline{\hashTableEmbedding}_{i}$, the nearest per-vertex feature $\vertFeat_{i\jKNN^{\ast}}$, the query point tangent coordinates $\queryPoint_{i\jKNN^{\ast}}$ of the nearest vertex applied with positional encoding, and the camera view direction with positional encoding. The tiny MLP consists of two hidden layers, where each hidden layer contains a Fully Connected layer with ReLU activation and 64 neurons. Please refer to Sec.3.3 and Fig.3 for more pipeline details. For the positional encoding, we use $8$ frequency bands on the query point tangent coordinates, and $4$ frequency bands on the camera view direction.

\subsection{Warp Field MLP}
The warp field MLP $\mathcal{F}_{\mathcal{E}}$ described in Sec.3.3 and \cref{sec:warp} consists of a backbone and three output branches. The backbone contains $5$ hidden layers, where each layer has $128$ neurons. Then, we append three branches, each is a $2$-layer MLP with $128$ neurons width, for regressing the three outputs described in \cref{sec:warp}: pure log-quaternion of the $3$D rotation (\ie, SO(3)) $\mathbf{R}$, rotation center $\mathbf{c}^{rot}$, and $3$D translation $\mathbf{t}$. ReLU activation is used in all layers except the output layers. We adapt a coarse-to-fine positional encoding strategy as used in Nerfies~\cite{park2021nerfies} on the query point coordinates before feeding into the MLP $\mathcal{F}_{\mathcal{E}}$ for better training stability. We start with $0$ frequency bands and increase to $6$ linearly after $80$k training iterations.

\section{Training and Testing Details}
\label{sec:train_details}

To obtain a consistent 3D world space, we normalize the 3DMM meshes with their neck poses to align the head in 3D space.
During training, we use a hierarchical sampling strategy as in \cite{mildenhall2021nerf}, where we use $32$ coarse and $32$ fine sample points per ray. During testing, we obtain a union occupancy grid for all training expressions, and run ray marching on those valid voxels to achieve efficient rendering. 
To ensure stable training, we enable 3D warping fields after 5k iterations. During training, we use the Adam optimizer with $\beta_{1}=0.9$, $\beta_{2}=0.999$. In each mini-batch, we random sample $256$ rays from $8$ images (\ie $2048$ rays in total) and set the learning rates to: (1) $10^{-4}$ for the warping field MLP and exponentially decay to $10^{-5}$ after $400$k. (2) $5 * 10^{-4}$ for other neural networks and exponentially decay to $5 * 10^{-5}$ after $400$k. We train the model with $400$k iterations for each subject.

\section{Data Settings}
\label{sec:data}

In \cref{tab:data}, we show more details on our data statistics over 10 subjects.

\begin{table}[!h]
\small
\centering
\setlength{\tabcolsep}{0.8em}
\begin{tabular}{l|c|c|c}
\toprule
    & \makecell{Number of \\ Train Frames} & \makecell{Number of \\ Test Frames} & Resolution \\
\midrule
    subject0 & $1560$ & $434$ & ($512$, $402$) \\
    subject1 & $1480$ & $740$ & ($512$, $422$) \\
    subject2 & $1440$ & $603$ & ($512$, $380$) \\
    subject3 & $1360$ & $564$ & ($512$, $368$) \\
    subject4 & $1450$ & $304$ & ($512$, $398$) \\
    subject5 & $2655$ & $595$ & ($512$, $372$) \\
    subject6 & $1818$ & $696$ & ($512$, $452$) \\
    subject7 & $3912$ & $817$ & ($512$, $512$) \\
    subject8 & $2656$ & $898$ & ($512$, $344$) \\
    subject9 & $2049$ & $351$ & ($512$, $512$) \\
\bottomrule
\end{tabular}
% \vspace{1mm}
\caption{Data statistics over 10 subjects.}
\label{tab:data}
\end{table}

\section{Additional Experiments}
\label{sec:results}

\subsection{Additional Results}

\begin{figure*}[htp!]
    \centering
    \includegraphics[width=\textwidth]{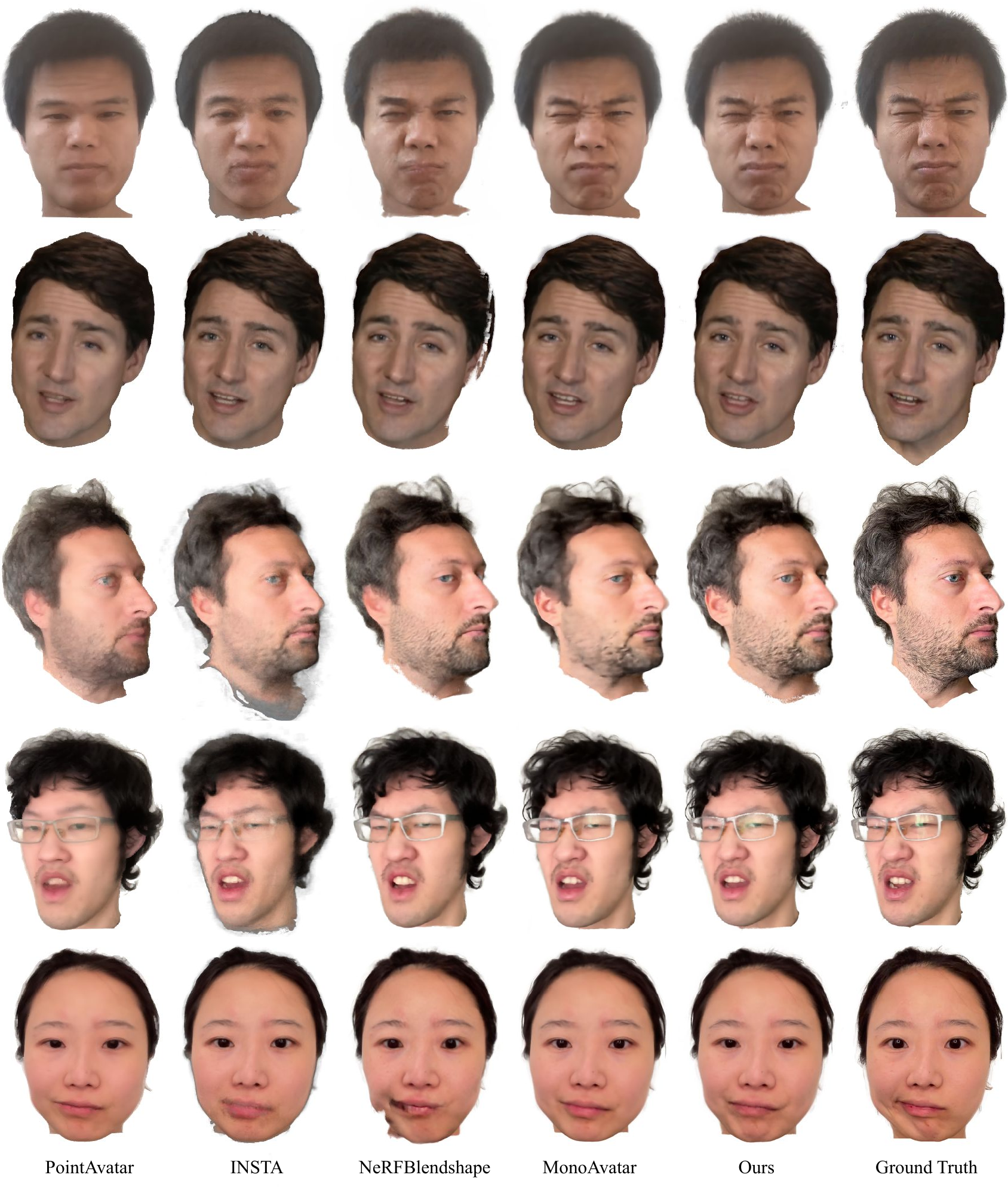}
    \caption{Comparisons on the rendering quality to previous state-of-the-art methods. From left to right, each column contains the images of: 1) PointAvatar~\cite{zheng2023pointavatar}, 2) INSTA~\cite{zielonka2023instant}, 3) NeRFBlendshape~\cite{Gao2022nerfblendshape}, 4) MonoAvatar~\cite{bai2023learning}, 5) Ours, 6) Ground Truth. Our method faithfully reconstructs the personalized expressions and high-frequency details, achieving one of the best rendering quality with real-time rendering speed.}
    \label{fig:vs_sota_1}
\end{figure*}

\begin{figure*}[htp!]
    \centering
    \includegraphics[width=\textwidth]{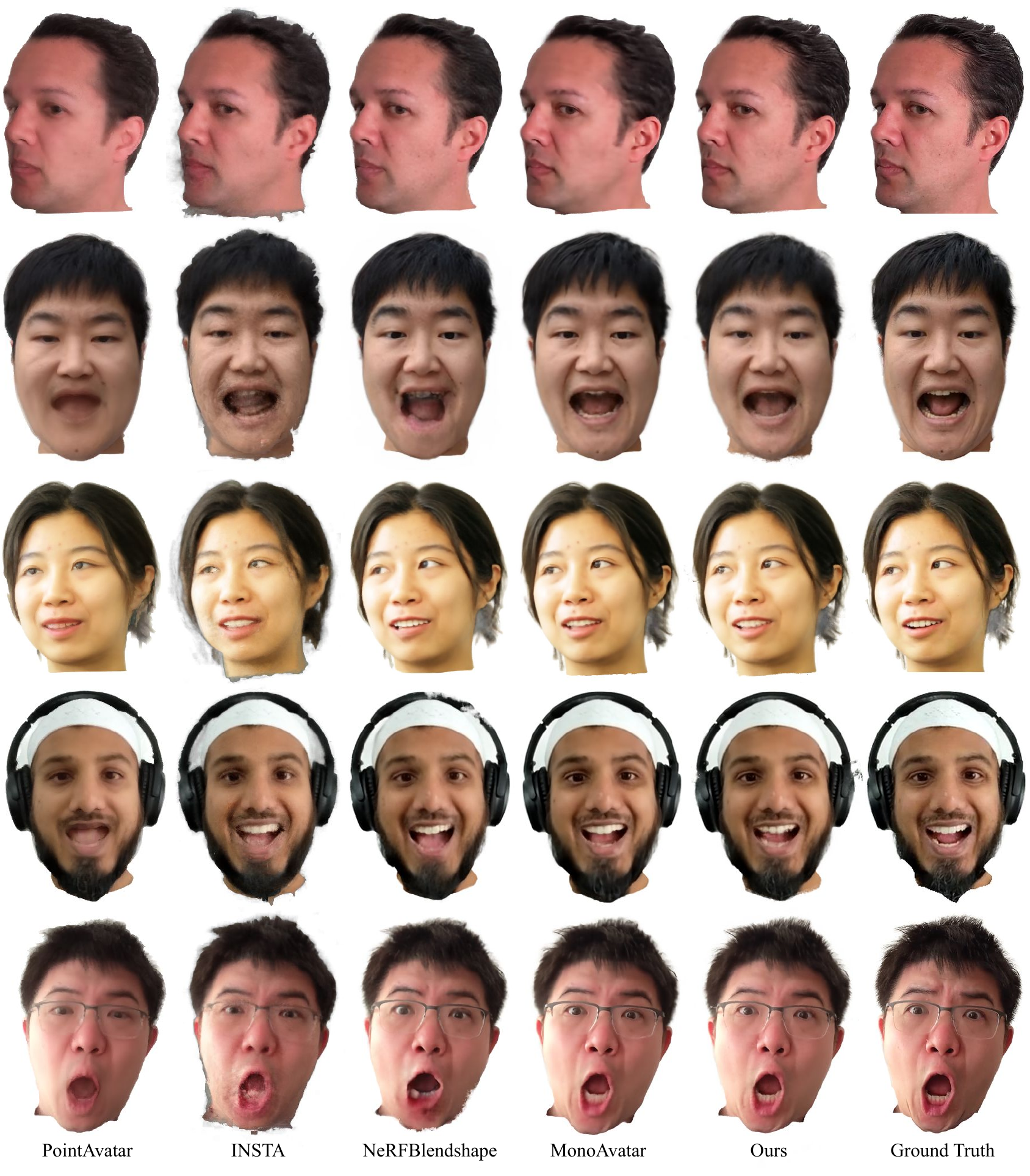}
    \caption{Comparisons on the rendering quality to previous state-of-the-art methods. From left to right, each column contains the images of: 1) PointAvatar~\cite{zheng2023pointavatar}, 2) INSTA~\cite{zielonka2023instant}, 3) NeRFBlendshape~\cite{Gao2022nerfblendshape}, 4) MonoAvatar~\cite{bai2023learning}, 5) Ours, 6) Ground Truth. Our method faithfully reconstructs the personalized expressions and high-frequency details, achieving one of the best rendering quality with real-time rendering speed.}
    \label{fig:vs_sota_2}
\end{figure*}

In this section, we provide additional experimental results and comparisons with prior state-of-the-art methods. Please see \cref{fig:vs_sota_1} and \cref{fig:vs_sota_2} for qualitative results and the accompanying \textbf{supplementary webpage} for video results.
% webpage: https://drive.google.com/file/d/1joG-PBNpfH7PjziW-k-ulSbT1_rPXFNo/view?usp=drive_link&resourcekey=0-QwYeK5CxX7GapiAt2sERMA

In \cref{fig:vs_sota_1} and \cref{fig:vs_sota_2}, we show more image results comparing with prior state-of-the-art approaches. PointAvatar~\cite{zheng2023pointavatar} and INSTA~\cite{zielonka2023instant} give overall inferior renderings than ours due to their limited model capacities in capturing static (\eg, glasses, hairs) and dynamic (\eg, expression-dependent deformations and wrinkles) avatar details. NeRFBlendshape~\cite{Gao2022nerfblendshape} produces less stable results, leading to severe artifacts around mouth and obvious floaters on avatar boundaries. MonoAvatar~\cite{bai2023learning} stably generates high quality renderings and animations, but is much slower than our method (\ie, $0.5$ FPS vs. $35.9 $ FPS) and slightly smoother on some details, for example, hairs, teeth and wrinkles. Our method overall achieves one of the best rendering quality, . Please refer to \cref{sec:limit} for more discussions on our limitations.

In the accompanying \textbf{supplementary webpage}, we include video results on various subjects with side-by-side comparisons to prior state-of-the-art methods, including PointAvatar~\cite{zheng2023pointavatar}, INSTA~\cite{zielonka2023instant}, NeRFBlendshape~\cite{Gao2022nerfblendshape}, and MonoAvatar~\cite{bai2023learning}. The videos show that our method is able to produce high-quality renderings while maintaining real-time speed.
% In addition, we also include video results of driving our trained avatars by the same subject under different capturing conditions than the original training sequences (\ie, novel conditions of lighting, appearance, and accessories, demonstrating the practical applicability of our method.

\subsection{Comparisons to More Works}
Here, we provide comparisons to more state-of-the-art methods that deliver relatively fast solutions for (partially) volumetric head avatar, including AvatarMAV~\cite{xu2023avatarmav} and LatentAvatar~\cite{xu2023latentavatar}. AvatarMAV~\cite{xu2023avatarmav} represents the head avatar by feature grid blendshapes to achieve fast training. LatentAvatar~\cite{xu2023latentavatar} learns a neural expression latent space instead of using 3DMM expression codes, and generate triplanes from this expression latent space. The triplane is rendered into a low resolution feature map, which is then used to synthesis the output RGB images via a 2D CNN.

As shown in \cref{tab:more_sota}, our method is able to achieve comparable rendering quality with these SOTA approaches, while supporting real-time rendering simultaneously. Note that LatentAvatar~\cite{xu2023latentavatar} uses heavy CNNs to directly synthesis output images, which leads to good sharpness (shown by LPIPS) but temporally 3D inconsistent high-frequency details. Also, directly synthesis image with CNN is an orthogonal direction to our method, which can also be appended to our pipeline.

\begin{table}[h]
% \vspace{-10pt}
\centering
% \footnotesize
\small
\setlength{\tabcolsep}{0.5em}
\begin{tabular}{l|ccc|c}
\hline
 & { LPIPS} & { SSIM} & { PSNR} & { Mean FPS} \\
\hline
% \multicolumn{5}{c}{\footnotesize \textit{Full Training Data}} \\
% \hline

AvatarMAV~\cite{xu2023avatarmav}    & 0.128 & 0.792 & 23.51 & 2 \\

LatentAvatar~\cite{xu2023latentavatar}    & 0.092 & 0.763 & 21.94 & 16 \\

% StyleAvatar & 0.087 & 0.678 & 21.17 & 29\\

Ours    & 0.100 & 0.795 & 22.77 & 35.9 \\

\hline

\end{tabular}
\caption{Quantitative comparisons with more state-of-the-art approaches. Our method achieves comparable rendering quality, while supports real-time rendering with a $512 \times 512$ resolution.}
\label{tab:more_sota}
% \vspace{-10pt}
\end{table}

\subsection{Ablation on Discarding Hash Tables}
Here, we investigate a new ablation setting \textit{No Hash + UV CNN}, where we discard all hash tables while keeping other parts unchanged. In this way, our model decodes the neural radiance field thoroughly from the vertex-attached features as in MonoAvatar~\cite{bai2023learning} but with a much smaller MLP for fast rendering. This gives the following results: $0.164$ / $0.759$ / $21.57$ for LPIPS / SSIM / PSNR, which are largely inferior than our full model \textit{Ours}. This indicates that the local hash tables are important for boosting the model capacity to achieve photorealism rendering quality.

\subsection{Geometry Visualization and Analysis}
\begin{figure*}[htp!]
    \centering
    \includegraphics[width=\textwidth]{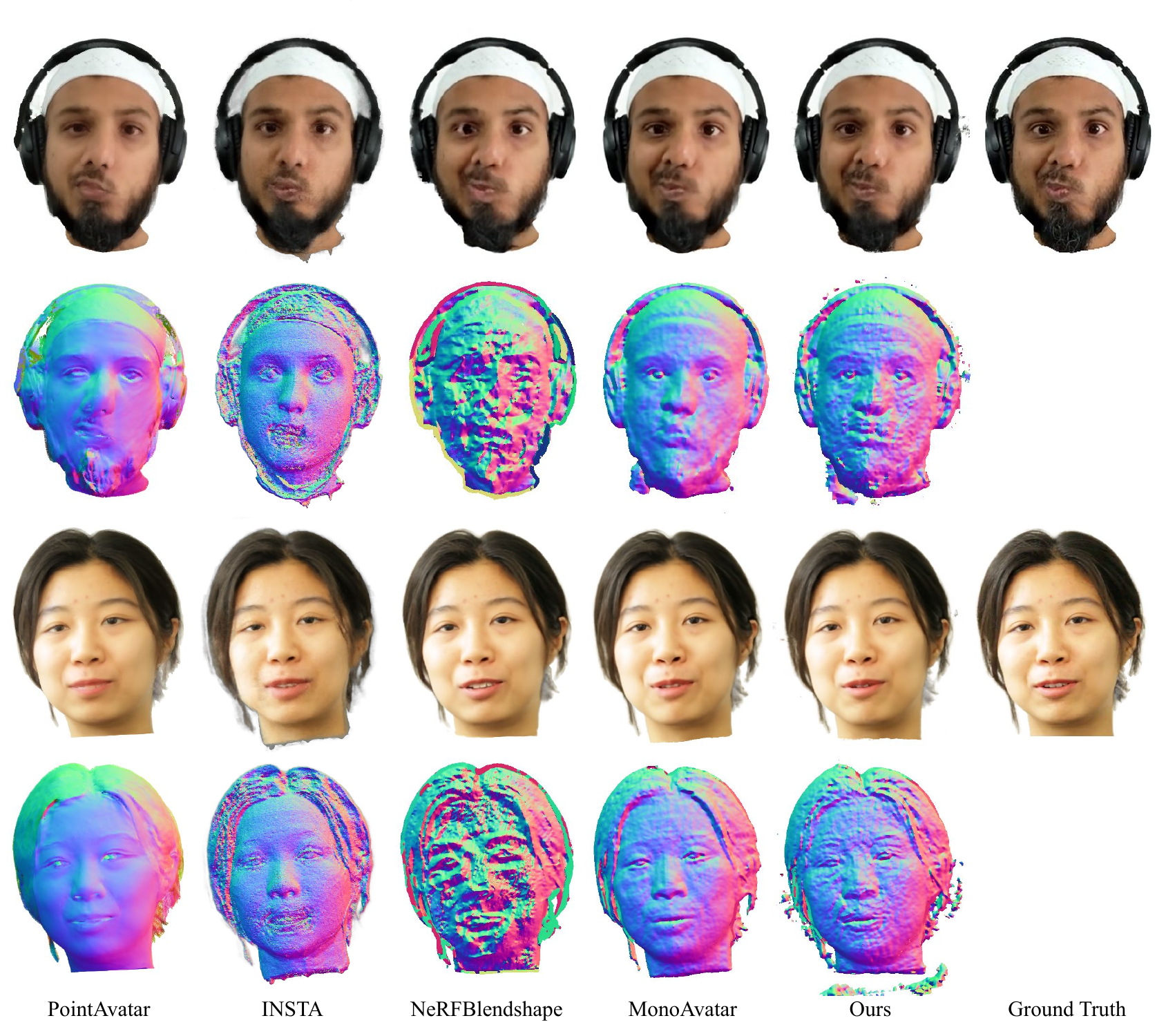}
    \caption{Visualization of normal maps and compare to previous state-of-the-art methods. From left to right, each column contains the images of: 1) PointAvatar~\cite{zheng2023pointavatar}, 2) INSTA~\cite{zielonka2023instant}, 3) NeRFBlendshape~\cite{Gao2022nerfblendshape}, 4) MonoAvatar~\cite{bai2023learning}, 5) Ours, 6) Ground Truth. Our method estimates reasonable geometries, while achieving one of the best rendering quality with real-time rendering speed.}
    \label{fig:normals_sota}
\end{figure*}

For the purpose of a comprehensive system analysis, we visualize the resulting geometry (as normal maps) of our avatars and compare with prior state-of-the-art approaches. \cref{fig:normals_sota} shows the normal map visualization. PointAvatar~\cite{zheng2023pointavatar} gives smooth geometry estimation thank to their relighting formulation. But their renderings are overall blurrier than other methods. INSTA~\cite{zielonka2023instant} generates geometries closing to the 3DMM meshes since they regularize the NeRF depth to the rasterized 3DMM depth on face region, which also leads to incorrect shapes for beard. Moreover, their rendered images are also suffered from unsatisfying quality. Despite NeRFBlendshape~\cite{Gao2022nerfblendshape} gives relatively good renderings, their estimated geometry is very noisy, presumably because that they do not leverage the 3DMM mesh as a shape prior. MonoAvatar~\cite{bai2023learning} gives both decent geometries and renderings, but is one order of magnitudes slower than real-time speed. Our method gives reasonable geometries that are slightly noisier than MonoAvatar, but supports real-time rendering speed while maintaining decent image quality.

\section{Limitations}
\label{sec:limit}

\begin{figure}[htp!]
    \centering
    \includegraphics[width=\linewidth]{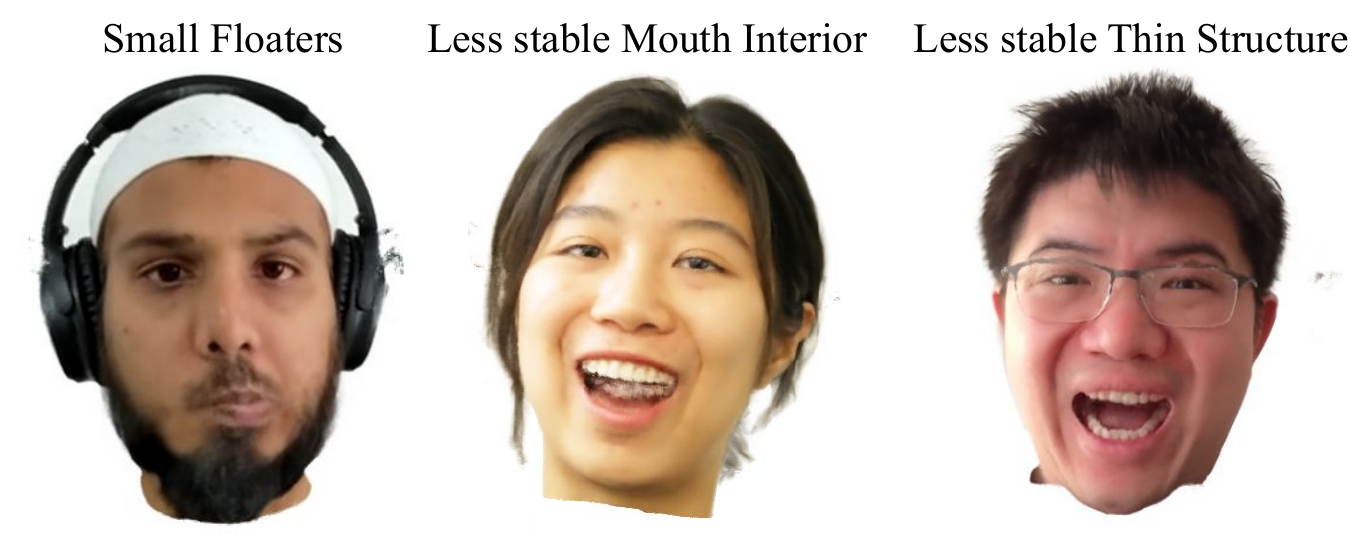}
    \caption{Due to the high flexibility of hash table embeddings and poor tracking on mouth interiors, we observe minor artifacts of our method, including small floaters, less stable mouth interiors and thin structures.}
    \label{fig:limit}
\end{figure}

Comparing with the state-of-the-art high quality MonoAvatar~\cite{bai2023learning}, our method is facing some quality trade-offs. On the positive side, our method captures more high frequency details such as hairs, teeth, and wrinkles thank to the high flexibility of hash table embeddings. However, this also introduces slightly more floaters than MonoAvatar~\cite{bai2023learning} (\cref{fig:limit}), which is a common issue for methods based on instant NGPs~\cite{mueller2022instant}. Presumably due to the same reason as well as poor tracking, we also observe a slightly less stale performance around the mouth interior regions and thin structures such as glasses frames. Further improving the robustness and stability without hurting quality and speed is an interesting future direction to explore.

\end{document}